# Creation of Digital Test Form for Prepress Department

Jaswinder Singh Dilawari
PhD, Research Scholar, Pacific University,
Udaipur, Rajasthan, India

Dr. Ravinder Khanna
Pricipal, Sachdeva Engineering College for Girls
Mohali, Punjab, India

*Abstract:* The main problem in colour management in prepress department is lack of availability of literature on colour management and knowledge gap between prepress department and press department. So a digital test from has been created by Adobe Photoshop to analyse the ICC profile and to create a new profile and this analysed data is used to study about various grey scale of RGB and CMYK images. That helps in conversion of image from RGB to CMYK in prepress department.

## I. INTRODUCTION

In the prepress process before printing, there is always need to produce image accurately and as per requirement of printing department. For this ICC has made a common colour profile and every printing process either develops their own printer profile considering ICC profile as reference or hire consultants. A biggest problem in creating own profiles is lack of literature available in market about colour management. Printers/prepress houses have difficulties in adjusting specific parameter settings in the profile, due to insufficient color management skills. However, if a flatbed scanner can precisely scan each color patch colorimetric correctly, then a scanner profile (ICC) would not be necessary and each color which is incorrectly scanned according to its colorimetric value will need a color correction when being converted from the source profile (scanner) to the destination profile (RGB color working space). The input image handled by prepress department is in the RGB mode (Red, Green, Blue) and to print it has to be converted to CMYK mode (Cyan, Magenta, Yellow, Black)[1][2]. This color conversion is today done by ICC-profiles. The profiles contain information about separation, black start, black width, total ink coverage. GCR (Gray Component Replacement) and UCR (Under Color Removal) are the two main color separation techniques used to control the amounts of black, cyan, magenta and yellow needed to produce the different tones. Since black ink can replace equal amounts of cyan, magenta and yellow. To produce a similar tone, UCR and GCR replace equal amounts of cyan, magenta and yellow in neutral tones. GCR also replaces some CMY colors in tertiary colors.[3] Before converting the image from RGB to CMYK first of file color profiles have to be understood. For this a digital test form has to be created using the most common software used in market- Adobe Photoshop. This digital test form will remove remedy of availability of literature in creating own profile. Many printers used external consultants for their color management and color profiles are created based on the information provided by them. So a specification is required to improve the communication between external consultant and printer as well as prepress and press department.

## II. CHARACTERIZATION OF INPUT DEVICES IN TERMS OF LUMINANCE AND CHROMINANCE

Each scanning device has to be characterized to common ICC color profile. To check their characterization a high Chroma image is used and tested using three different IT8.7/2 test targets [4]. Besides the established IT8-targets from the major color chart vendors a new IT8-target was created for the tests. The four test charts are named A, B, C and D in the study. Reference color values such as lightness and chroma coordinates were read from the test targets. A spectrophotometer was used for the readings. The fourth test chart will be created by using the analysis data of three test charts A, and C. These test charts will be compared with ISO-standard ISO 12641-1997 and their profile spaces are also compared and analyzed. The result of this analysis is used to create the new test chart and its effect on color gamut is also studied.

*A. Creation of New IT-8 Test Chart*

The new test chart created follows the ISO standard LCH (ISO 12641-1997). The scanner target consists of a total of 264 colors, as shown in Figure 1. The target design is a uniform mapping and is defined in detail in the ANSI standard IT8.7/2 for reflection material (ISO 12641-1997).

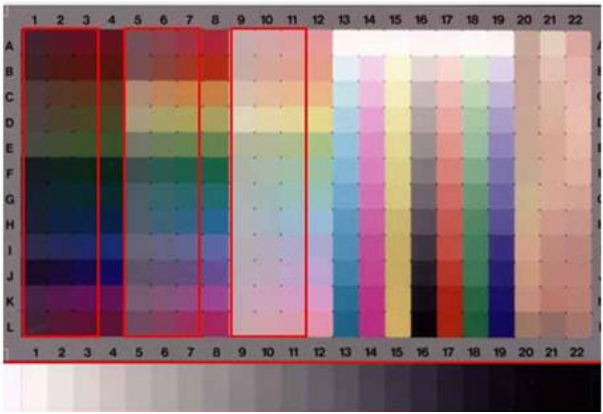

Figure 1: The scanner target consists of a total of 264 colors. The red frames show the standardized values.

As is shown in the above figure three luminance levels are defined with 12 hue angles. Each hue angle for each luminance level has four chrominance values and the highest chrominance value is unalterable [5]. A further 84 patches provide additional tone scales which are not defined by any ISO-standard. Seven tone scales are defined for the colors cyan, magenta, yellow, red, green and blue (no ISO standard defined). Each tone scale is built-up in twelve steps starting from the lowest chrominance value and keeping the hue angle stable. Each vendor has defined an optimal tone scale for their own specific output media [6]. The last three columns are vendor specific. Here the vendor manufacturing a target was allowed to add any feature they deemed worthwhile. By keeping above specifications in mind following IT-8 test chart is developed as shown in figure 2.

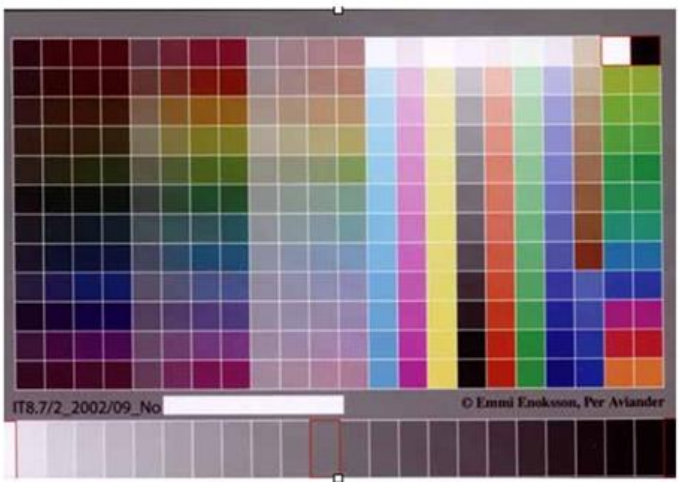

Figure 2: The customized IT.8 target for scanners

## III PROBLEM IN CALIBRATING DIFFERENT SCANNERS

The analysis of problem in calibration of scanners involves two different studies at different span of time.

### A. First Study

The first study was performed in 2000 when ICC-profiles were used by only a minority of Swedish printers. Color separation, at that time, was performed directly in image scanners or in imaging applications (i.e. Adobe Photoshop) using color look up tables. A total of 120 companies, both printers with prepress departments and dedicated prepress houses, participated in the study. The companies are all located in Sweden, with an even geographical spread throughout the nation. The printers and prepress houses were also chosen on the basis of the size of the company, but only companies with two or more employees were included in the survey. Semi-structured interviews were conducted with prepress representatives, normally by telephone or by e-mail. Ten company visits were made. A number of questions concerning the different separation techniques were asked in order to be able to assess the general level of competence [7].

### B. Outcome of Study

After the study it has been observed that there is poor communication between prepress and press department, lack of knowledge of image separation, no proper image separation. Moreover there is need for guidelines written in understandable form. It has been observed that only 20 % of printers and prepress house has a good knowledge of conversion of image from RGB to CMYK.

### C. Second Study

The second study was performed in 2003. Eighty sheet-fed offset printers and thirty four newspaper printers, evenly geographically spread over Sweden, participated in this study. Companies with only one employee were not included. As in the first study, semi-structured interviews were conducted with prepress representatives for each printer or prepress house either through a visit or by e-mail. A structured web questionnaire was also used. The questions asked concerned the use, creation and implementation of ICC-profiles [8]. Approximately 50 percent of the printers/prepress houses participating in this study were also involved in the first study. In order to verify the findings and clarify the results, nine independent color consultants were contacted and interviewed.

### C. Outcome of Study

The second study showed that a majority (70%) of the commercial printers nationwide in Sweden are using ICC-profiles for color reproduction, particularly in the newspaper industry (83%). The majority of the participants in the survey felt that there was a lack of communication or non-existing communication in all process directions. There is normally no dedicated time for quality meetings [9]. The newspapers have a better know-how than commercial printers concerning color management. Few companies set a strategy for their color management implementation and they therefore may not use the consultant in the right way. Terminology confusion is common in the graphic arts industry. The study shows that many pre-press staff members use the terms incorrectly or mix them up. The survey indicates that external consultants play an important role in the creation of ICC-profiles.

*D. Where is the problem?*

The problem lies at ground level. The calibration is difficult because of lack of knowledge and communication gap in various departments. The context of knowledge lacking lies in less availability of literature available on color management instructions which could help printers to better understand the technology [10]. The communication problem was due to a lack of a common language, due mainly to the different backgrounds and experiences of the people involved. The creation of new ICC profile is not an easy process. Main problems a respondent faces are:

1) Device calibration

2) Misunderstood profiling set-up options

3) Lack of understanding of the profiling process

4) Inappropriate test target

5) Inappropriate profiling software

The quality can be improved by removing these problems during building of color management software.

IV. CREATION OF DIGITAL TEST FORM

Generally printing personnel don't bother about the maintenance of production quality because of lack of time and pressure of completion of work on time. To read manuals and thereby learn more about certain software has been shown to be a poor alternative, as manuals are often written in a difficult way, and often in a foreign language. So graphic art industry needs a tool which can be used by the user by himself [11][12][13]. User can use its application and setting for betterment of image quality and later on can use these setting for better production. For this digital test forms are created using Adobe Photoshop. The created test form can provide information to the user about many settings in the profile. The test form helps to show the differences between the settings already in the RGB color mode and to avoid misunderstandings after printing. The layout of the test form facilitates practical understanding by showing the result of a color conversion from RGB to CMYK using a profile.

The digital test form gives information about:

• ICC-profiles in MacOS and Windows

• Different color gamuts

• RGB Gray balance

• Rendering intents

• Gamut warning

• Separation

• CMYK gray balance

• Chroma shift

• Gamut mapping

• Skin tones

• Total Ink Coverage

The image in figure2 can be analyzed in context to change in profile with the help of digital form factor.

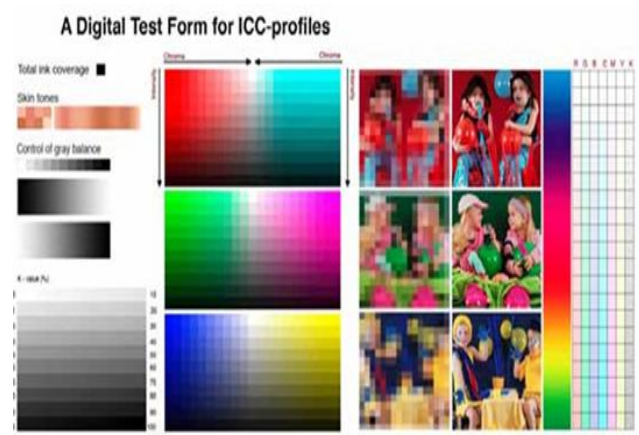

Figure 3: The digital test form for the evaluation of ICC-profiles

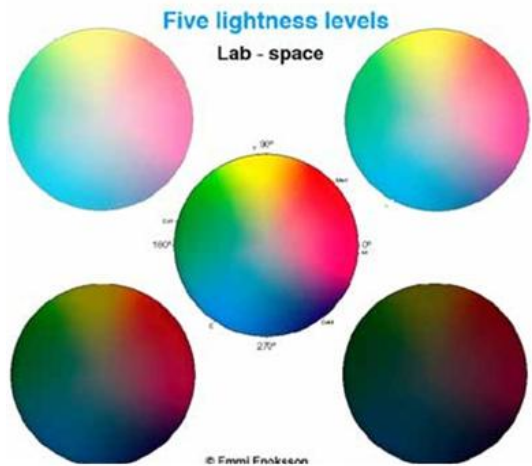

Figure 4: The lightness circles in five levels for evaluation. To better understand how the settings affect the result the user can see the changes in two directions - the vertical direction and the horizontal direction (the lightness circles on the digital test form).

V. COMMUNICATION BETWEEN PROCESSES

Before designing a new ICC profile specifications should be clearly specified. An understanding of these specifications will facilitate internal communication within a printing company, and also between printing companies and external consultants creating ICC-systems [14]. It has been observed that none of the printing company is having clear specifications for the construction of ICC profiles. In printing process every process (prepress, press, post press) should communicate together and to communicate properly input and output specifications of these processes should be understood properly. An improved communication will give a better process understanding and thereby a better production quality [15][16]. For better communication between processes these five points should be considered:

*1)General demands/specifications*

This part of the list contains objectives, implementation and specifications. Before a profile-based production can be considered, comprehensive objectives need to be set. The purpose of the process change must be explained to the personnel directly involved in the production process. General information about profile implementation needs to be given [17]. To avoid common misunderstandings and improve internal communication, written process instructions should be followed. Each process should be definedand described, with regard to responsibility and demands.

*2) Test form specifications*

This part describes responsibility distribution according to the creation and content of the test form.

*3) RIP (Raster Image Processor) specifications*

This part of the communication list describes initial demands - linearization and setting of the RIP.

*4) Output profile specifications*

This part describes responsibility distribution according to settings in the profile.

*5) Printing specifications*

This part describes initial demands, general facts and standard demands.

Using above methods if specifications are clearly understood then communication between processes will be proper and will result in proper color management. Each step in a color management set-up must be documented so that a later profiling update can be established with the same set-up [18][19][20]. The communication list deals with specification demands which are of importance in the development of profiles and different responsibility distributions in the development of these profiles. Two scenarios are described: the first situation is when the printing company creates its own profiles without the involvement of an external consultant, and the second scenario describes the situation when the printing company needs external help to create the profiles.

V. CONCLUSION

After talking with various printing presses and their correspondent, it has been observed that the lack of literature available for the color management is the main problem. When different scanning devices or prepree and press departement have to communicate then also lack of communiaction is there because of no knowledge of color management. So after studying problem digital forms have been created using Adobe Photoshope which can give information about chrominance value and hue angles in an image. By studying this digital form the communication gap between various departemnt can be coverd. Moreover the problem of literature available in different languages is also overcome because digital test form represents color test chart in user readable format. Moreover digital test form gives information about RGB gray balance and CMYK gray balance. So the conversion of RGB to CMYK in prepress departement is eased by digital test form.

**Jaswinder Singh Dilawari** is working as an Associate Professor ,Geeta Engineering College, Panipat, Haryana ,India .He has teaching experience of 12 years .His area of interest includes Computer Graphics, Computer Architecture ,Software Engineering ,Fuzzy Logic and Artificial Intelligence .He is life member of Indian Society for Technical Education (ISTE)

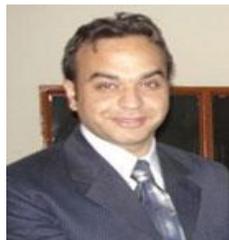

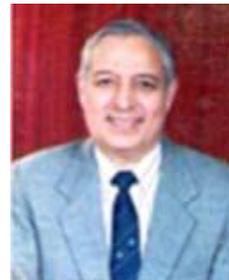

Born in 1948, Dr. Ravinder Khanna Graduated in Electrical Engineering from Indian Institute of Technology(IIT) Dehli in 1970 and Completed his Masters and Ph.D degree in Electronics and Communications Engineering from the same Institute in 1981 and 1990respectively. He worked as an Electronics Engineer in Indian Defence Forces for 24Years where he was involved in teaching, research and project management of some of the high tech weapon systems. Since 1996 he has full time Switched to academics. he has worked in many premiere technical institute in india and abroad. Currently he is the Principal of Sachdeva Engineering College for Girls, Mohali, Punjab (India).He is active in the general area of Computer Networks, Image Processing and Natural Language Processing.